\documentclass[runningheads]{llncs}
\usepackage{graphicx}
\usepackage{amsmath,amssymb} 
\usepackage{ruler}
\usepackage{color}
\usepackage{multirow}

\usepackage{cases}
\usepackage{bm}
\usepackage{algorithm}
\usepackage{algorithmic}
\usepackage{cases}
\usepackage{subfigure}
\usepackage{enumerate}
\usepackage{etoolbox}
\usepackage{boxedminipage} 

\DeclareMathOperator*{\argmax}{arg\,max}

\newcommand{\MyMapTemplatePrefix}[4]{\expandafter#1\csname#3#4\endcsname{#2{#4}}}
\newcommand{\MyMapTemplatePrefixNew}[5]{\expandafter#1\csname#4#5\endcsname{#2{#3{#5}}}}
\forcsvlist{\MyMapTemplatePrefix {\def} {\mathbf} {}} {A,B,C,D,E,F,G,H,I,J,K,L,M,N,O,P,Q,R,S,T,U,V,W,X,Y,Z}
\forcsvlist{\MyMapTemplatePrefix {\def} {\mathbf} {}}
{1,0}
\forcsvlist{\MyMapTemplatePrefix {\def} {\mathbf} {}}
{a,b,c,d,e,f,g,h,i,j,k,l,m,n,o,p,q,r,s,t,u,v,w,x,y,z,1,0}
\forcsvlist{\MyMapTemplatePrefix {\def} {\widetilde} {wt}} {A,B,C,D,E,F,G,H,I,J,K,L,M,N,O,P,Q,R,S,T,U,V,W,X,Y,Z}
\forcsvlist{\MyMapTemplatePrefix {\def} {\widetilde} {wt}} {a,b,c,d,e,f,g,h,i,j,k,l,m,n,o,p,q,r,s,t,u,v,w,x,y,z} 
\forcsvlist{\MyMapTemplatePrefixNew {\def} {\widetilde}{\mathbf} {tb}} {A,B,C,D,E,F,G,H,I,J,K,L,M,N,O,P,Q,R,S,T,U,V,W,X,Y,Z}
\forcsvlist{\MyMapTemplatePrefixNew {\def} {\ddot}{\mathbf} {db}} {A,B,C,D,E,F,G,H,I,J,K,L,M,N,O,P,Q,R,S,T,U,V,W,X,Y,Z}
\forcsvlist{\MyMapTemplatePrefixNew {\def} {\check}{\mathbf} {cb}} {A,B,C,D,E,F,G,H,I,J,K,L,M,N,O,P,Q,R,S,T,U,V,W,X,Y,Z}
\forcsvlist{\MyMapTemplatePrefixNew {\def} {\widetilde}{\mathbf} {tb}} {a,b,c,d,e,f,g,h,i,j,k,l,m,n,o,p,q,r,s,t,u,v,w,x,y,z}
\forcsvlist{\MyMapTemplatePrefixNew {\def} {\ddot}{\mathbf} {db}} {a,b,c,d,e,f,g,h,i,j,k,l,m,n,o,p,q,r,s,t,u,v,w,x,y,z}
\forcsvlist{\MyMapTemplatePrefixNew {\def} {\check}{\mathbf} {cb}} {a,b,c,d,e,f,g,h,i,j,k,l,m,n,o,p,q,r,s,t,u,v,w,x,y,z}
\forcsvlist{\MyMapTemplatePrefix {\def} {\widehat} {wh}} {A,B,C,D,E,F,G,H,I,J,K,L,M,N,O,P,Q,R,S,T,U,V,W,X,Y,Z}
\forcsvlist{\MyMapTemplatePrefix {\def} {\widehat} {wh}} {a,b,c,d,e,f,g,h,i,j,k,l,m,n,o,p,q,r,s,t,u,v,w,x,y,z}
\forcsvlist{\MyMapTemplatePrefixNew {\def} {\widehat}{\mathbf} {hb}} {A,B,C,D,E,F,G,H,I,J,K,L,M,N,O,P,Q,R,S,T,U,V,W,X,Y,Z}
\forcsvlist{\MyMapTemplatePrefixNew {\def} {\widehat}{\mathbf} {hb}} {a,b,c,d,e,f,g,h,i,j,k,l,m,n,o,p,q,r,s,t,u,v,w,x,y,z}
\forcsvlist{\MyMapTemplatePrefix {\def} {\mathcal}{mc}} {A,B,C,D,E,F,G,H,I,J,K,L,M,N,O,P,Q,R,S,T,U,V,W,X,Y,Z}
\forcsvlist{\MyMapTemplatePrefix {\def} {\mathbb} {mb}} {A,B,C,D,E,F,G,H,I,J,K,L,M,N,O,P,Q,R,S,T,U,V,W,X,Y,Z}
\forcsvlist{\MyMapTemplatePrefix {\DeclareMathOperator} {} {} } {tr,diag,sgn}
 
\def\etc{\emph{etc.}}
\def\ie{\emph{i.e.}} 
 \def\eg{\emph{e.g.}}

\newcommand{\tabincell}[2]{\begin{tabular}{@{}#1@{}}#2\end{tabular}}

\makeatletter
\def\blfootnote{\gdef\@thefnmark{}\@footnotetext}
\makeatother

\begin{document}
\def\ACCV16SubNumber{} 

\title{Recurrent Regression for Face Recognition} 

\author{Yang Li, Wenming Zheng, Zhen Cui}
\institute{Key Laboratory of Child Development and Learning Science of Ministry of
Education, Research Center for Learning Science, Southeast University, Nanjing,
Jiangsu Province 210096, China\\\{yang\_li; wenming\_zheng; zhen.cui\}@seu.edu.cn}

\maketitle

\begin{abstract}
To address the sequential changes of images including poses, in this paper we propose a recurrent regression neural network(RRNN) framework to unify two classic tasks of cross-pose face recognition on still images and video-based face recognition. To imitate the changes of images, we explicitly construct the potential dependencies of sequential images so as to regularize the final learning model. By performing progressive transforms for sequentially adjacent images, RRNN can adaptively memorize and forget the information that benefits for the final classification. For face recognition of still images, given any one image with any one pose, we recurrently predict the images with its sequential poses to expect to capture some useful information of others poses. For video-based face recognition, the recurrent regression takes one entire sequence rather than one image as its input. We verify RRNN in static face dataset MultiPIE and face video dataset YouTube Celebrities(YTC). The comprehensive experimental results demonstrate the effectiveness of the proposed RRNN method.
\end{abstract}

\section{Introduction}

Face recognition is a classic topic in past decades and now still attracting much attention in the field of computer vision and pattern recognition. With the rapid development of electric equipment techniques, more and more face images can be easily captured in the wild, especially video sequences from cameras of surveillance or cell phones. Therefore, video or image set based face recognition becomes more important to flatter those real-world applications and also a popular topic in face analysis more recently. As face images captured from the unconstrained conditions are usually with complex appearance variations in poses, expressions, illuminations, \etc, the existing face recognition algorithms still suffer from a severe challenge in fulfilling a real applications to large-scale data scenes, although the current deep learning techniques have made a great progress on the unconstrained small face dataset, \eg, the recent success of deep learning methods on Labeled Faces in the Wild (LFW)~\cite{huang2014labeled}.

In the task of face recognition, however, we cannot bypass this question of pose variations, which has been extensively studied and explored in past decades, and has not been well-solved yet. The involved methods may be divided into 3D \cite{blanz2003face,drira20133d,asthana2011fully} and 2D methods \cite{kan2014stacked,tan2010enhanced,naseem2010linear,chen2014cross}. Since pose variations are basically caused by 3D rigid motions of face, 3D methods are more intuitive for pose generation. But 3D methods usually need some 3D data or recovery of 3D model from 2D data which is not a trivial thing. Moreover, the inverse transform from 3D model to 2D space is sensitive to facial appearance variations. In contrast to 3D model, due to decreasing one degree of freedom, 2D methods usually attempt to learn some transforms across poses, including linear models~\cite{asthana2013robust} or non-linear models~\cite{kan2014stacked}. Due to its simplicity, 2D model has been widely used to deal with cross-pose face recognition with a comparable performance with 3D model. However, in many real scenes of face image sets, \eg, face video sequences, the changes of poses may be regarded as a nearly-continuous stream of motions, while the existing methods usually neglect or do not make full use of this prior. Moreover, the pose variation is not the only factor between different images even for the same subject, which involves other complex factors.

To address the sequential changes of images including poses, in this paper we propose a recurrent regression neural network(RRNN) framework to unify two classic tasks of cross-poses face recognition on still images and video based face recognition. To imitate the changes of images, we explicitly construct the potential dependencies of sequential images so as to regularize the final learning model. By continuously transferring information from sequentially adjacent images, RRNN can adaptively memorize and forget the information that benefits for the final classification. For face recognition of still images, given any one image with any one pose, we recurrently predict the images with its sequential poses to expect to capture some useful information of others poses, under the supervision of known pose sequences. For video-based face recognition, we regress on one entire sequence rather than one image in contrast to still images based face recognition. By repetitively regularizing the relationship of adjacent frames, we can obtain more robust representation of face video sequences under the supervised case. To verify our methods, we conduct two experiments, one is face recognition across poses on Multi-PIE dataset~\cite{gross2007cmu} and the other is video-based face recognition on YouTube Celebrities (YTC) dataset~\cite{kim2008face}, which is collected from internet sources in the wild. Experimental results demonstrate that the proposed RRNN is very effective and consistently superior to the current state-of-the-art methods.


\section{Related Work}

\subsection{Face Recognition across Poses}

With the development of 3D camera technology, several researches try to solve face recognition problem with 3D face images. 3D face images can solve the problem that the distance between two certain parts of face vary in different poses, and 3D face images contain more information about the face such as the height of the facial features. The current 3D technologies include 3D images processing (captured by 3D camera) and 3D recovery(transformed from 2D to 3D images(2D$\rightarrow$3D)). For example, \cite{drira20133d} proposed a novel geometric framework for analyzing 3D faces. It represents facial surfaces by radial curves emanating from the nose tips and uses elastic shape analysis of these curves to develop a Riemannian framework for analyzing shapes of full facial surfaces. \cite{asthana2011fully} proposed a 3D pose normalization method that is completely automatic and leverages the accurate 2D facial feature points found by the system. \cite{li2012morphable} proposed a novel method, named Morphable Displacement Field (MDF), using a virtual view to match the pose image.

3D technology has been confirmed to own a good performance in face recognition~\cite{drira20133d,asthana2011fully}. However, 3D data is hard to get in some unconstrained scenes, and the 2D to 3D algorithms are still needed to be research.

Meanwhile, there have existed a lot of algorithms for traditional 2D face image recognition. For pose variation face recognition,~\cite{sharma2011bypassing} linearly mapped images in different modalities to a common linear subspace in which they are highly correlated.~\cite{sharma2012generalized} presented a general multi-view feature extraction by learning a common discriminative subspace, in which pose variation is minimized.~\cite{kan2014stacked} proposed a stacked progressive auto-encoders network, which changes the larger poses to near frontal pose.

\subsection{Video-based Face Recognition}

Video-based face recognition is generally studied for three scenes, namely Still-Video, Video-Still and Video-Video~\cite{lu2015multi}. Still-Video face recognition searches the static image in a video. It is always used to find a man in a video when given only a face image of him. On the contrary, Video-Still face recognition matches a video of a man in a lot of images. Video-Video inquires the clip of a man's video given in a lot of videos. Video-based face recognition is different from face image recognition. Under normal circumstances, the faces captured by a camera are always affected by the environment around seriously and sometimes have low quality, \eg  large angle pose, blur, low resolution and complex illumination. Consequently, video-based face recognition is more challenging than still image face recognition. Especially, we noticed that head in videos usually swings, which makes faces have different poses in each frame.

Benefiting from the good performance of 2D face recognition technology, video-based face recognition causes several researchers' attention.~\cite{hadid2009manifold} proposed a novel approach based on manifold learning to solve the problem of video-based face recognition in which both training and test sets are video sequences.~\cite{chen2012dictionary} introduced the concept of video-dictionaries for face recognition, which generalizes the work in sparse representation and dictionaries for faces in still images.

\subsection{Recurrent Neural Networks}

\label{RNN}
Recently, RNNs have achieved a big success in pattern recognition area. For example, RNNs show good performance on speech recognition~\cite{graves2013speech} and hand-writing recognition~\cite{graves2009offline}.

In traditional neural networks, all inputs(and outputs) are independent with each other. However it neglects the inputs that are associated with each other. Sentence prediction is an example. If we want to get the next word in the sentence, we must associate it with the words before. Because each words in a sentence are not independent with others. Traditional RNNs learn complex temporal dynamics by mapping input sequences to a sequence of hidden states.

Recurrent Neural Network(RNN) is designed to process sequence data. The outputs of this layer have a relationship with the outputs before. That makes it have a memory to remember the information before.

\section{Recurrent Regression}
\label{RR}

In this section, we first provide an overview introduction on the proposed recurrent neural network framework, then two cases of face recognition based on cross-pose and video are further modeled.

\subsection{The Model}

\begin{figure}
\centering
\includegraphics[width=0.9\columnwidth]{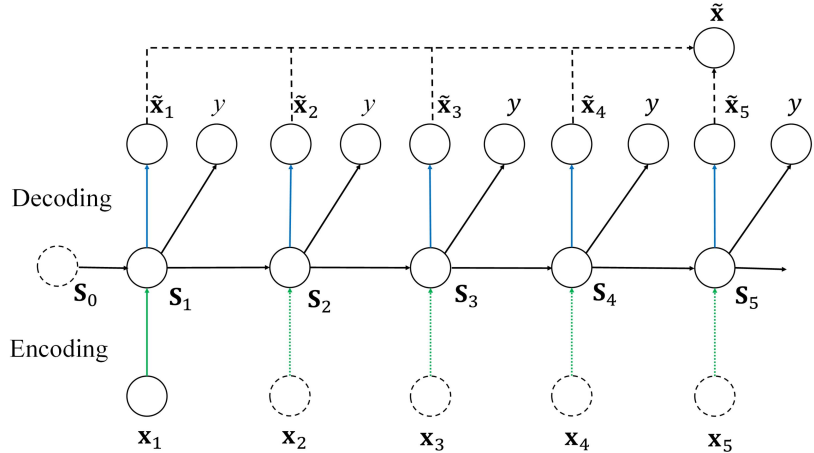}
\caption{An illustration of the overall framework of our proposed recurrent regression neural network.}
\label{fig2}
\end{figure}

The overall framework of RRNN is shown in Fig.~\ref{fig2}. To make full use of various of appearance models for still image or video based face recognition, we explicitly build a recurrent regression model to transform the current input into other appearance spaces, in which we seek for some effective components to compensate mismatching appearance variations of face images. Given an input $\x_i$, we encode it into a latent state $\S_i$ and then decode it to one virtual output $\tbx_i$, which may come from the other space spanned by some other appearance characteristics we expect. The encoder-decoder models a dynamic changing process between the input and the expected output, and may be further stacked layer by layer to represent a sequence, \ie, a process of recurrent encoding-decoding. In the task of face recognition, in order to enhance the model discriminability, the identification of subjects can be combined into this model as a joint learning. Besides, to reduce error-drifting of all decodings, we impose a total error constraint on the sum of all outputs to explicitly smooth the entire output sequence. Concretely, we formulate recurrent regression on a sequence of appearance variations from three aspects:
\begin{enumerate}
\item[(1)] Recurrent encoding-decoding. Let $\{\x_0,\x_1,\cdots,\x_{t-1},\x_t,\cdots\}$ denote the input sequence states, then the corresponding hidden states $\{\S_0,\cdots,\S_t,\cdots$\} after encoding can be written as
\begin{eqnarray}
\mathbf{S}_t=\sigma(\mathbf{U}\mathbf{x}_t+\mathbf{W}\mathbf{S}_{t-1}+\mathbf{b}_1),
\end{eqnarray}
where \textbf{U}, \textbf{V} and \textbf{W} are linear transform weight matrices, $\textbf{b}_1$ is the bias term, and $\sigma$ is a nonlinear transform activation function, \eg, the tanh function used in this paper. Note that, the encoding also depends on the previous hidden state $\S_{t-1}$ partly besides the current input because previous historic information may bring external beneficial information for the next representation. Further, we decode each hidden state $\S_t$ into its specified output $\tbx_t$ we expected, \ie,
\begin{eqnarray}
\tbx_t=\sigma(\mathbf{V}\mathbf{S}_t+\mathbf{b}_2),\qquad t = 1,2,\cdots,
\end{eqnarray}
where $\V$ is the decoding matrix and $\textbf{b}_2$ is the bias. Consequently, this objective function is to minimizing all reconstruction errors, \ie,
\begin{eqnarray}
f_1=\sum_{t = 1,2,\cdots} \|\hbx_t-\tbx_t\|_F^2,\label{eqn:f1}
\end{eqnarray}
where $\hbx_t$ is the ground-truth of the next state in the sequence.

\item[(2)] Sequence reconstruction. To further characterize the globality of the decoding on a sequence, we force some statistic properties of all outputs to close to be an expected state $\tbx$, \ie, minimizing the following objective function,
\begin{eqnarray}
f_2 = \|\tbx -g(\x_1,\x_2,\cdots,\x_t,\cdots)\|_F^2,\qquad t = 1,2,\cdots,
\end{eqnarray}
where $g$ is the statistic function on a sequence, such as first-order statistics. Different from the separate error in Eq.~(\ref{eqn:f1}), this term is used to collaborate all reconstruction units to reduce error-propagation, which can further improve the performance according to our observation from the following experiments.
\item[(3)] Discriminative prediction. Like most supervised models, we may add a supervision term into the network so as to enhance the model discriminability. Concretely, we use logistic regression on the transformed hidden states, \ie,
\begin{eqnarray}
&& f_3 = -\sum_t \log(P(\wty_t=i|\S_t,\G,\b)),\\
&&\wty_t = \argmax P(y_t=i|\S_t,\G,\b),\\
&&P(y_t=i|\S_t,\G,\b)= \frac{\exp(\mathbf{G}\mathbf{S}_t+\mathbf{b}_3)}{\sum_k \exp(\mathbf{G}\mathbf{S}_k+\mathbf{b}_3)},
\end{eqnarray}
where the variables $\G$ and $\b_3$ are respectively the transform matrix and the bias. Note that here the supervision information is directly imposed on the hidden states rather than the decoding output $\tbx_t$. The reasons are two folds: i) the reconstruction in each decoding unit is not perfect, where the errors might reduce the discriminative capability especially when accumulatively propagated along the sequent network; ii) it can implicitly transit some identification information to the reconstruction stage and thus reduce the direct influence on decoding targets due to the large sematic gap between reconstruction targets and labels.
\end{enumerate}

Thus, the overall objective function can be defined as
\begin{eqnarray}
\label{erro}
\min\quad f_1 + \alpha f_2 + \beta f_3,
\end{eqnarray}
where $\alpha$ and $\beta$ are the balance parameters corresponding to the sequence reconstruction term and the supervision term.

\subsection{Still Images across Poses}
\label{Sec:PRNN:pose}

For still images with different poses, the poses can be sorted in a sequence according to the continuity of pose changing. In the training stage, the image sequence along pose changes may be easily captured from cameras, and thus can be used in the proposed recurrent regression network. However, in the classic task of cross-pose face recognition based on still images, only one image is usually provided for testing. So we have to flatter this model by converting one image into a virtual sequence.

Given any one image $\x_1$ with one pose, we augment it into a sequence stream by using repeatedly copy, \ie, $\{\textbf{x}_1, \textbf{x}_1,...,\textbf{x}_1,\cdots\}$, which is pretended to be the input sequence. For the decoding outputs, we expect to predict those images of other poses. In order to utilize gradual changes of poses, we construct the decoding output sequence as the adjacent pose stream $\{\textbf{x}_2, \textbf{x}_3,...,\textbf{x}_{t+1},\cdots\}$, \ie, the next adjacent pose is its decoding output. In this way, recurrent encoding-decoding can realize the function that transforms the input pose to the target pose we expect. For the sequence reconstruction term, we use the mean values of predicted state as first-order statistics, and then make it close to the mean of pose streams. This term
could collaborate all reconstruction poses to reduce error-propagation and also take advantage of the global information.

\subsection{Video Sequences}

Different from the above case in Section \ref{Sec:PRNN:pose}, for video-based face recognition, the input sequence is explicitly known in the testing stage. For RRNN, thus the input sequence consists of all frames of a video sequence, \ie, $\{\textbf{x}_1, \textbf{x}_2,...,\textbf{x}_t,\cdots\}$, where $\textbf{x}_t$ is the $t$-th frame of the sequence. Instead of the use of next frames, we use the mean value of all frames as the decoding outputs we expect, \ie, $\{\overline{\x},\overline{\x},\cdots,\overline{\x},\cdots \}$, where $\overline{\x}=\frac{1}{n}\sum_{t=1}^n \x_t$. The main reason is that our aim is to classify each sequence rather than predict next frames. If we use next frames as the decoding outputs, we could capture more motion information in the encoding-decoding process, which does not refine the subject information yet. Under this constraint of mean prediction, the sequence reconstruction term will play an unimportant role in the final performance also as observed from our experiments, due to the nearly-common optimization target.





\section{Experiments}

\subsection{Setting up}

\label{Experiment result}

In this section, we evaluate our proposed RRNN on two widely used face dataset, one is the cross-pose face dataset MultiPIE~\cite{gross2007cmu}, and the other is the video dataset YouTube celebreties(YTC)~\cite{kim2008face}. As Convolutional Neural Network(CNN) can extract much robust features according to the recent researches, so in this experiment we employ CNN features to represent the images to feed into RRNN as inputs. Concretely, we directly employ the released training model of VGGFACE~\cite{Parkhi15} network to extract face features, where the output of 2622 dimension on the layer `fc8' are used as the features of each image. Of course, the CNN model can be concatenated with our RPNN for an end-to-end neural network. Considering a small scale training samples, we only use it to extract features to verify our idea. Without fine tuning on the network parameters, we simply set the number of hidden units to 5000 as default value in the following experiments.

\subsection{Face Recognition across Poses}

MultiPIE dataset contains 337 people with face images of different poses, illumination and expressions. Each person has 7 poses
from $45^\circ$ to $-45^\circ$ with $15^\circ$ interval, where $0^\circ$ means the frontal pose. Following the same experiment configuration to~\cite{kan2014stacked}, we choose the first 200 subjects(subject ID from 1 to 200) as the training set, totally 4207 face images. The rest 137 subjects are used as the testing set, totally 1879 face images. Inside the testing set, we take one frontal pose face image from each subject, totally 137 frontal pose face images as the gallery set. The rest 1742 face images are used as probe set.

As face images with $0^\circ$ pose are easier to be recognized according to human cognition, we convert each pose to the frontal pose by using the gradual changing strategy. For example, given an image of $-45^\circ$ pose, we expect the decoding sequence to be faces with $\{-30^\circ,-15^\circ,0^\circ\}$ poses. However, images of different poses will have a regression sequence with different lengthes. To handle this problem, we pad the frontal pose into those short sequences so as to generate the encoding sequences with equal length. In order to identify the end of front pose in the testing stage, we externally extend the sequence by adding the frontal pose at the end of each sequence, where we expect the front pose will not be changed in the looped regression model as the terminate state.

As the training set and the gallery set do not share the same label information, we do not use the discriminative term in RRNN, which in fact directly verifies our idea of recurrent regression itself. Given a testing sample, we take the mean value of hidden states as the regressed features, and employ K Nearest Neighbor(KNN) algorithm to classify it. For PRNN, we set the balance parameter $\alpha=0.1$ in Eq.~(\ref{erro}). Table~\ref{tab1} shows the results.

In order to test the performance of our RRNN, we compare it with the state-of-the-art methods, consisting of 3D and 2D technologies. For 3D technologies, we compare the two methods of Asthana11~\cite{asthana2011fully} and MDF~\cite{li2012morphable}. For 2D technologies, we compare those classic models including PLS~\cite{sharma2011bypassing}, CCA~\cite{hotelling1936relations}, GMA~\cite{sharma2012generalized}, DAE~\cite{bengio2009learning} and SPAE~\cite{kan2014stacked}, where SPAE achieves the current best performance on this protocol. To verify the effectiveness of the regression model in principle, we also conduct the experiment VGG+KNN. The comparison results are reported in  Table~\ref{tab1}, we can have two main observations from it:
\begin{enumerate}
\item[1)] From the penultimate line of this table, VGG~\cite{Parkhi15}+KNN achieves the best performance compared the existing state-of-the-art methods, even the recently proposed deep learning method SPAE. It again indicates that CNN can benefit face recognition more than those raw/hand-crafted features.
\item[2)] Although CNN features are robust enough, RRNN can further improve the performance by using the prior of pose changing. Our RRNN performs better than all of the compared algorithms in all poses especially in $-45^\circ$, $-30^\circ$ and $45^\circ$. Compared to SPAE which results are recently released, the average improvement is about 4 percents.
\end{enumerate}

\begin{table}[!t]
\centering
\caption{The classification results on MultiPIE dataset.}
\renewcommand{\arraystretch}{1.3}
\begin{tabular}{|c|c|c|c|c|c|c|c|}
\hline
\multirow{2}{*}{Methods} & \multicolumn{7}{c|}{Probe Pose} \\ \cline{2-8}
{} & $-45^\circ$ & $-30^\circ$ & $-15^\circ$ & $+15^\circ$ & $+30^\circ$ & $+45^\circ$ & Average \\ \hline \hline
Asthana11 & 74.1\% & 91.0\% & 95.7\% & 95.7\% & 89.5\% & 74.8\% & 86.8\% \\ \hline
MDF & 78.7\% & 94.0\% & 99.0\% & 98.7\% & 92.2\% & 81.8\% & 90.7\% \\ \hline
PLS & 51.1\% & 76.9\% & 88.3 \% & 88.3\% & 78.5 \% & 56.5\% & 73.3\% \\ \hline
CCA & 53.3\% & 74.2\% & 90.0\% & 90.0\% & 85.5\% & 48.2\% & 73.5\% \\ \hline
GMA & 75.0\% & 74.5\% & 82.7\% & 92.6\% & 87.5\% & 65.2\% & 79.6\% \\ \hline
DAE & 69.9\% & 81.2\% & 91.0\% & 91.9\% & 86.5\% & 74.3\% & 82.5\% \\ \hline
SPAE & 84.9\% & 92.6\% & 96.3\% & 95.7\% & 94.3\% & 84.4\% & 91.4\% \\ \hline
VGG+KNN & 83.0\% & 94.9\% & 98.6\% & 97.9\% & 94.6\% & 85.8\% & 92.5\% \\ \hline
\textbf{RRNN} & 90.4\% & 96.9\% & 98.6\% & 97.6\% & 96.2\% & 91.3\% & 95.2\% \\ \hline
\end{tabular}
\label{tab1}
\end{table}

\subsection{Video-Video Face Recognition}

YouTube celebreties(YTC)~\cite{kim2008face} dataset contains 1910 face videos of 47 people. These videos are with large variation of pose, illumination and expression. As compression ratio of most videos is very high, the quality of faces in video are usually very poor,  especially including some factors of blur, low-resolution, fast motions, \etc Furthermore, the number of video frames ranges from 7 to 400.

As described in~\cite{lu2015multi}, we detect the faces in YTC videos and align them into $20\times20$. Following the similar protocol to ~\cite{lu2015multi}, we randomly choose a video of each session of each subject for training, and choose 2 videos from the rest videos for each session for testing. As total 3 sessions, thus 3 samples of each subject are used for training and 6 samples for testing. Ten trials are randomly conducted so as to cover all samples. The average accuracy of ten trials are used the final result. In this training, to reduce the computation cost of each sequence and increase the training sequences, we cut each video to several clips of 10 frames. Due to the shared labels of training set and testing set, we use the discriminative model (\ie, logistic regression) to predict the classification score.


Here we compare RRNN with several state-of-the-art algorithms, including MSM~\cite{yamaguchi1998face}, DCC~\cite{kim2006learning}, MMD~\cite{wang2008manifold}, MDA~\cite{wang2009manifold}, AHISD~\cite{cevikalp2010face}, CHISD~\cite{cevikalp2010face}, SANP~\cite{hu2011sparse}, CDL~\cite{wang2012covariance}, DFRV~\cite{chen2012dictionary}, LMKML~\cite{lu2013image}, SSDML~\cite{zhu2013point}, SFDL~\cite{lu2014simultaneous} and MMDML~\cite{lu2015multi}. Their mean accuracies are reported in Table~\ref{tab2}. These methods fall into the category of subspace based or metric based methods. It is apparently RRNN gets the best performance compared with all the other algorithms. What is more, the improvement is up to 7.2$\%$. This huge improvement indicates RRNN can well-model video sequence.

\begin{table}[!t]
\centering
\caption{Average classification result and standard deviation on YTC dataset.}
\renewcommand{\arraystretch}{1.3}
\begin{tabular}{|c|c|c|c|c|c|c|c|}
\hline
Methods & MSM & DCC &MMD & MDA & AHISD & CHISD & SANP \\ \hline
YTC & 61.7$\pm$4.3 & 65.8$\pm$4.5 & 67.7$\pm$3.8 & 68.1$\pm$4.3 & 66.5$\pm$4.5 & 67.4$\pm$4.7 & 68.3$\pm$5.2 \\ \hline
Year & 1989 &2006 & 2008 & 2009 & 2010 & 2010 & 2011 \\ \hline \hline

Methods & CDL & DFRV& LMKML& SSDML & SFDL & MDML& RRNN \\ \hline
YTC & 69.7$\pm$4.5 & 74.5$\pm$4.5 & 75.2$\pm$3.9 & 74.3$\pm$4.5 & 75.7$\pm$3.4 & 78.5$\pm$2.8 & 86.6$\pm$1.1 \\ \hline
Year & 2012 & 2012 & 2013 & 2013 & 2014 & 2015 & \\ \hline

\end{tabular}
\label{tab2}
\end{table}

\subsection{Discussion}

\subsubsection{The effectiveness of terms in the objective function.} As described in the objective function Eq.~(\ref{erro}), there are two related parameters $\alpha$ and $\beta$, which respectively constrain the sequence reconstruction and label prediction.

According to the above analysis, in the task of face recognition across poses, we do not use the label prediction term, \ie, the case of $\beta=0$, which means the term $f_3$ can not be calculated. With recurrent encoding-decoding $f_1$, the recognition accuracy is up to 94.7\%. Furthermore, by adding sequence reconstruction $f_2$, the performance is further improved up to 95.2\% when $\alpha=0.1$.

In the task of video-based face recognition, we discard the sequence reconstruction term $f_2$ due to the common target with the encoding-decoding process as analyzed above. By adding the label loss term $f_3$, we can promote the performance about 1 percent when $\beta=1$.

For $\alpha$ and $\beta$, we only tune their in the range $\{0.1,1\}$ without finer tuning. Although so, we find they benefit the final classification performance by introducing the learning of appearance variations.


\subsubsection{Cross-pose Analysis.} Although in the above experiments on MultiPIE the frontal pose is specified as the terminate state of the proposed recurrent regression model, we can directly obtain all cross-pose results based on this frontal pose model by selecting a pose as the gallery set and the rest poses as the probe set. Table~\ref{tab3} shows the results of different poses as gallery sets for face recognition across poses on MultiPIE. It is interesting to observe that the recognition result don't reach the best when $0^\circ$ images are used as the gallery set, which seems not match with our intuition. The reasons should come from two aspects: 1) Frontal faces in the gallery set are currently decoded along the time stream, and reconstruction errors are propagated with the evolution of states, which leads to a derivation from the ground-truth frontal faces for the decoding states. 2) According to the symmetry of faces, non-frontal faces can induce frontal faces to some extends as non-frontal faces contain more contour information than frontal faces.

\begin{table}[t]
\centering
\caption{Cross-pose results of our proposed RRNN on MultiPIE.}
\renewcommand{\arraystretch}{1.3}
\begin{tabular}{|c|c|c|c|c|c|c|c|c|c|}
\hline
\multicolumn{2}{|c|}{\multirow{2}{*}{}} & \multicolumn{8}{c|}{Probe Pose} \\ \cline{3-10}
\multicolumn{2}{|c|}{} & $-45^\circ$ & $-30^\circ$ & $-15^\circ$ & $0^\circ$ & $+15^\circ$ & $+30^\circ$ & $+45^\circ$ & Average \\ \hline
\multirow{8}{*}{\tabincell{c}{Gallery\\ Pose}}
& $-45^\circ$ & {-} & 0.9861 & 0.9861 & 0.9695 & 0.9686 & 0.9685 & 0.9709 & 0.9749 \\ \cline{2-10}
{} & $-30^\circ$ & 1.0000 & {-} & 1.0000 & 0.9926 & 0.9967 & 0.9966 & 0.9783 & 0.9940 \\ \cline{2-10}
{} & $-15^\circ$ & 0.9853 & 1.0000 & {-} & 1.0000 & 0.9967 & 1.0000 & 0.9674 & 0.9916 \\ \cline{2-10}
{} & $0^\circ$ & 0.8934 & 0.9765 & 0.9867 & {-} & 0.9833 & 0.9663 & 0.9058 & 0.9520 \\ \cline{2-10}
{} & $+15^\circ$ & 0.9632 & 1.0000 & 1.0000 & 1.0000 & {-} & 1.0000 & 0.9855 & 0.9915 \\ \cline{2-10}
{} & $+30^\circ$ & 0.9706 & 0.9966 & 0.9967 & 1.0000 & 1.0000 & {-} & 0.9964 & 0.9934 \\ \cline{2-10}
{} & $+45^\circ$ & 0.9851 & 0.9788 & 0.9823 & 0.9769 & 0.9823 & 0.9894 & {-} & 0.9825 \\ \hline
\end{tabular}
\label{tab3}
\end{table}

\section{Conclusion}

In this paper, we proposed a Recurrent Regression Neural Network(RRNN) to unify two classic face recognition tasks including cross-pose face recognition and video-based face recognition. In RRNN, three basic units are considered to model a potential sequence data. The first unit is the encoder-decoder, which is used to model sequential reconstruction. The second unit is to constrain the globality of the sequence. The final one is to utilize the discriminative label information. By properly choose the configuration for different tasks, we can benefit from these units. Experimental results strongly indicate our RRNN achieves the best recognition compared with those state-of-the-art methods.



\bibliographystyle{splncs}
\bibliography{accv2016submission}

\end{document}